*-REVIEW ARTICLE-*

# Superiorities of Deep Extreme Learning Machines against Convolutional Neural Networks

Gokhan Altan[1*], Yakup Kutlu[1]

[1] Department of Computer Engineering, Iskenderun Technical University, Iskenderun, Hatay, Turkey

**Abstract**

Deep Learning (DL) is a machine learning procedure for artificial intelligence that analyzes the input data in detail by increasing neuron sizes and number of the hidden layers. DL has a popularity with the common improvements on the graphical processing unit capabilities. Increasing number of the neuron sizes at each layer and hidden layers is directly related to the computation time and training speed of the classifier models. The classification parameters including neuron weights, output weights, and biases need to be optimized for obtaining an optimum model. Most of the popular DL algorithms require long training times for optimization of the parameters with feature learning progresses and back-propagated training procedures. Reducing the training time and providing a real-time decision system are the basic focus points of the novel approaches. Deep Extreme Learning machines (Deep ELM) classifier model is one of the fastest and effective way to meet fast classification problems. In this study, Deep ELM model, its superiorities and weaknesses are discussed, the problems that are more suitable for the classifiers against Convolutional neural network based DL algorithms.

---

[*] *Corresponding Author: Gokhan ALTAN, e-mail: gokhan.altan@iste.edu.tr*

**Introduction**

Deep Learning (DL) is a classification method that is a specific version of the artificial neural networks (ANN). The DL was applied to various types of disciplines including computer vision, object detection, semantic segmentation, diagnosis systems, time series analysis, image generating, and more. The ANN model has limitations on layer size, number of neurons depending of computation capability of the CPU. Hence, whereas the number of the ANN model increases, the classification parameters to optimize increase along with the model advancing. The DL has advantages of pre-defined parameters and using shared weights in training stages. It has been the easiest way of producing new ANN models using most effective framework background for last decades. The DL has effective algorithms that are group demanding on the unsupervised and supervised stages at the training phases. The DL is a two-step classifier. At the first stage, the classification parameters are defined without labels using an unsupervised algorithm including filtering, restricted Boltzmann machines, autoencoders, self-organizing maps, convolution, stage mapping and more. The pre-defined parameters are fed into the ANN model with similar structure of the unsupervised stage with the labels to optimize the classification parameters for minimum error using back-propagation algorithms including contrastive divergence, dropout, extreme learning machines (ELM) and more.

Convolutional neural networks (CNN) model is the most popular method in the DL. The CNN is utilized to handle for visual imagery including the computer vision and object detection solutions. It provides to collect deterministic features from the input images using low- and high-order features with specific filters. The convolution filter size and number of the filters enables to perform self-assessment on the images. The convolutional layer and extracting deterministic variable order features are called as feature learning for the DL stages. The CNN was utilized to perform object detection, time-series signals, diagnosis of the different diseases using biomedical signals, face recognition and similar disciplines. Lawrance et al (1997) proposed a face recognition model and compared the self-organizing map approach with the CNN. They also performed hybrid classifier models on face images from 40 subjects. They highlighted the efficiency of the hierarchical set of layers on CNN for facial recognition models. Li et al. (2015) analyzed the face images to extract a face recognition model. They structured a cascade architecture on CNN. Their model had focused on multiple resolutions using central processing unit (CPU) and graphical processing unit (GPU) supports. Kalchbrenner et al. (2014) analyzed the sentences using CNN. They performed an accurate natural language processing model. The CNN has increased the performance of semantic word modelling system about 25%. Ciresan et al. (2011) suggested a CNN model for handwritten character optimization. They achieved 0.27% error rate of classification on MNIST dataset. Their analysis is a path for the next character optimization techniques with the simplicity of the model. The biggest step on modeling the CNN on big image datasets was ImageNET progresses and the analysis on it. Krizhevsky et al. (2012) has suggested ImageNet with novel models and layer sizes for image classification models. The analyzed image dataset is large and effective enough as a big data problem.

Deep ELM classifier is an alternative method to the DL algorithms. It is a fast and high-generalized capability model against the DL algorithms. It is based on the autoencoder modeling approach that creates representations in a selective output size. At the first stage, autoencoder kernel generates different presentations of the input data, and transfers the output weights into the queued layers of the deep model (Ding et al., 2015). At the last stage, last generated representations are fully connected to the supervised ELM classifier with the class labels. Deep ELM model was performed to analyze time-series for modeling diagnosis systems, image classification, character optimization and object detection (Altan & Kutlu, 2018a).

The remaining of the study is organized as definition and specifications of the Deep ELM and CNN in detail, applications of the classifiers, superiorities and performance comparison on various disciplines in the material method. The comparisons are based on the classification accuracy rates and training time for the same datasets in the literature. The advantages of the Deep ELM and CNN algorithms are discussed at the results section.

**Methods**

Classifier models are explained. The specifications of the classifier models with optimization parameters and requirements of the classifier models are shared for the beginner researchers on the DL applications. Even though the CNN and Deep ELM models are similar classifier models with the pre-defining specifications and supervised stages, the mathematical solutions and the workloads of the algorithms are completely different from each other.

**Convolutional Neural Networks**

The CNN is a main and frequently used type of DL for image-based approaches. Advantages of the CNN are standing feature learning with convolutional processes with iterated filters and sizes, and extracting the most significant bits at a specified range on the image using pooling process. The feature-learning phase of the CNN stands for the unsupervised stage of the model. The extracted pooling inputs are fed into the fully connected neural network model. The general structure of the CNN is depicted in Figure 1.

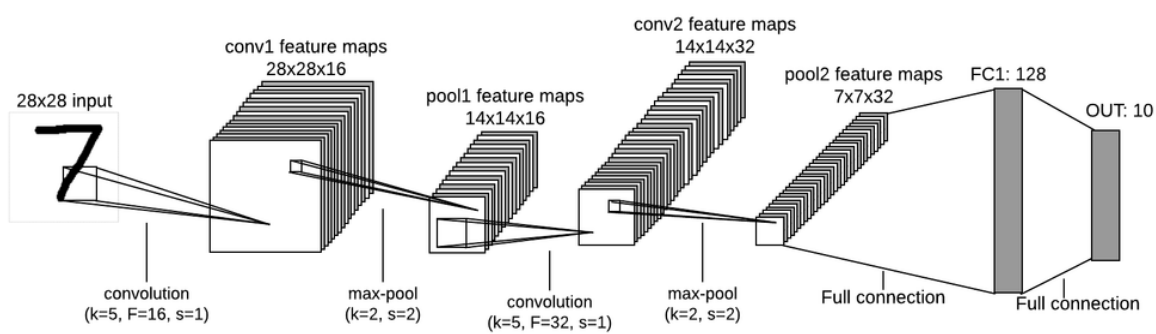

**Figure 1.** Structure of the CNN with two convolutional layers (Tensorflow, 2018)

The CNN considers angular edges with different edge-finding filters, transitions from dark to light, light to dark and calculates each one separately. The edges are usually detected in the first layers of a CNN. Whereas performing all of these convolutional calculations, there is a difference between the input size and the output size.

After convolutional layer, the calculation of the size difference between the input sign and the exit sign is possible. This is performed by adding extra pixels to the input matrix. This is called a padding. If the input matrix is n by n, the filter matrix (f by f) is the same size as the input matrix; (n + 2p-f + 1) * (n + 2p-f + 1) is applied. The value represented by id p is a matrix that is the pixel size added to the input matrix, the padding value. To determine this, p = (f-1) / 2 equation is used.

The stride stands for the weight matrix for the conversion operation will shift the filter in steps of one or more steps on the image. Stride is another parameter that affects the output size directly. It allows you to analyze low-level and high-level attributes on the entire image.

The max pooling method is usually used in the pooling layer. There are no parameters learned in this layer of the network. It decreases the height and width information by keeping the channel number of input matrix constant. Pooling layer performs decreasing image size for extracting distinctive features for smaller ones. At the fully connected layer, the bits of last convolved image is transferred as input values to the ANN.

**Deep Extreme Learning Machines**

ELM model is a feedforward neural networks structure with a single hidden layer. It is usually utilized as classifier, application of regression, clustering, decoding, encoding and feature learning. ELM does not require optimization for the classification parameters as different from ANN. The classification parameters including output weights, weight of hidden neurons are generated according to the input feature set and class labels (Kutlu et al., 2017). The generated output and input weights are never be tuned. It is a fast and stable algorithm for the classification stages. Hence, it is more stable and appropriate for real-time and instantaneous applications. Using only a single layer has limitations for detailed analysis of the input data.

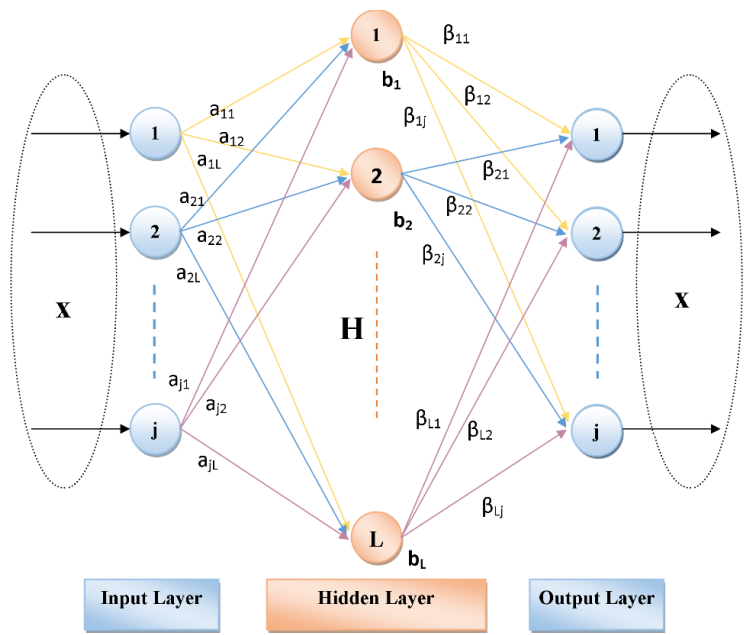

**Figure 2.** Structure of ELM Autoencoder for generating presentations for the Deep model

The theorem of ELM (Huang et al., 2006; Huang & Chen, 2007) is randomly stated input weights provides fast learning and the universal estimation ability using least mean square algorithm for defining outputs. $\beta$ is output weight matrix, $T$ stands for training data, $H$ is randomly generated hidden layer output matrix, $H^{\dagger}$ is the Moore-Penrose inverse of matrix of $H$.

$$\beta = H^{\dagger}T \tag{1}$$

Training process of the ELM is as follows:

$$\beta = H^T \left(\frac{1}{\lambda} + HH^T\right)^{-1} T \tag{2}$$

Deep ELM model is the enhanced type of the ELM. Deep ELM enables using multi-layer hidden nodes in the model with the ELM kernels. The autoencoder kernels are the basis of the transformation of the ELM into the Deep ELM models. An autoencoder is algorithm that generates the output weights in unsupervised ways effectively (Liou et al., 2014). The autoencoder generates a representation of input data with decreased or increased size of dimensionality models. The autoencoder algorithms are popular algorithms to enhance and accelerate DL classification stages (Altan & Kutlu 2018b). The ELM autoencoder gets the input matrix as also output matrix to generate the encoder vector (Figure 2). The encoder vector does not need optimization and carries deterministic and significant characteristic for input feature set. The transpose of calculated output weight ($\beta^T$) is used as the weight between two adjacent layers. At the last layer, the training progress is finalized with traditional ELM model in supervised ways.

**Conclusion**

The DL algorithms are effective ways for classification stages. Nevertheless, they requires optimization and feature learning stages that are time-consuming progresses. Especially, optimization of the classification parameters including output weights, learning rates, output functions, layer size, neuron size at each layer, convolution size and number of filtering channels, and more is a hard-to decide options for creating an accurate model for the disciplines. Reducing the number of the optimization parameters at training stage accelerates decision accuracy for a specific training range.

The CNN has an unchallengeable system performance on the different classification problems. Some studies are also performed ELM models to the convolutional features for achieving fast classification results. The Deep ELM has high and effective achievements on the same datasets in remarkable training time. The advantages of the Deep ELM is generalization performance, training speed and decoding application by the use of autoencoder models. It also has ability to analyze the input data in detail by using many hidden layers and autoencoder kernels. Each kernel and input with different sizes generates different presentations of the input feature set. The CNN has ability to extract low- and high-level features using convolutional layers. Nevertheless, using rectified linear unit activation function generates the sharp features from the images. CNN also utilized for assessing the time-series signal by converting them into the images. In our opinion, it is a disadvantage of the CNN, by an additional progress to the classification problems. For any signal, it is a handicap using non-standardized signals. The table below presents some studies on same problems.

**Table 1.** Training speed and classification performance comparison of CNN and Deep ELM

|  | Methods | Accuracy | Training Time |
|---|---|---|---|
| MNIST character recognition | Deep ELM | 99.14% | 281.37s (CPU) |
|  | CNN | 99.05% | 19 hours (GPU) |
| 3D Shape Classification | Deep ELM | 81.39% | 306.4s (CPU) |
|  | CNN | 77.32% | > 2 days (GPU) |
| Traffic sign recognition | Deep ELM | 99.56% | 209s (CPU) |
|  | CNN | 99.46% | >37 hours (GPU) |
|  | CNN + ELM | 99.48% | >5 hours (CPU) |

The CNN needs long time processes. The main improvements on DL routes the researchers to compose fast training methods. Deep ELM were proposed to advance the training capabilities for reducing time and generalization capabilities. Deep ELM model consists of autoencoder models, which are fast and effective models for generating different representations of the input data. Although Deep ELM uses CPU, it is much faster than CNN models. Anyso the use of GPU provides an advantage; training time is quite a noticeable difference between the models. The training time shows that, the Deep ELM model including ELM autoencoder kernels can be first choice to enhance the models into the real-time application even for CPU. Another alternative is integrating the Deep ELM model into the supervised stage of the CNN.

The ELM, which is a single layer neural network model, is formed to the many hidden layer using autoencoder kernels unsupervised ways. The proposed ELM autoencoder kernels including lower-upper triangularization (Altan and Kutlu, 2018a) and Hessenberg decomposition (Altan and Kutlu, 2018c) have improved the generalization capability of the DL approaches. The Deep ELM provides reducing fast training with simplest mathematical solutions. Deep ELM is an adaptable method that can be integrated even for supervised and unsupervised stages of the classifiers. It is possible to integrate the ELM autoencoder kernels into the filtering progress to generate the significant presentations of the input data. Using max pooling for the Deep ELM structures would provide extracting heavy low- and high- level features from the intended sizes of the input signals. The Deep ELM is also more compatible for the time-series to analyze the long-term signals using encoding specifications.